\documentclass[conference]{IEEEtran}
\IEEEoverridecommandlockouts
\usepackage{cite}
\usepackage{amsmath,amssymb,amsfonts}
\usepackage{algorithmic}
\usepackage{graphicx}
\usepackage{textcomp}
\usepackage{xcolor}
\usepackage{comment}
\usepackage{csquotes}
\usepackage{array}
\usepackage[most]{tcolorbox}
\usepackage{hyperref}
\usepackage{adjustbox}
\usepackage{booktabs}
\usepackage{tabularx}
\usepackage{ragged2e}
\usepackage{arydshln}
\usepackage{siunitx}

\newcolumntype{C}[1]{>{\centering\arraybackslash}m{#1}}

\def\BibTeX{{\rm B\kern-.05em{\sc i\kern-.025em b}\kern-.08em
    T\kern-.1667em\lower.7ex\hbox{E}\kern-.125emX}}
\begin{document}

\title{Large Language Models Produce Responses Perceived to be Empathic

\thanks{This project has benefitted from the Microsoft AI, Cognition, and the Economy (AICE) research program.
  This work was also partially supported by NSF grant IIS-2107524.}}


\makeatletter
\newcommand{\linebreakand}{%
  \end{@IEEEauthorhalign}
  \hfill\mbox{}\par
  \mbox{}\hfill\begin{@IEEEauthorhalign}
}
\makeatother

 \author{
     \IEEEauthorblockN{Yoon Kyung Lee}
     \IEEEauthorblockA{\textit{Department of Psychology} \\
     \textit{The University of Texas at Austin} \\
     \texttt{yklee@utexas.edu}
     }
 \and
     \IEEEauthorblockN{Jina Suh}
     \IEEEauthorblockA{\textit{Microsoft Research} \\
     \texttt{jinsuh@microsoft.com}
     }
\and
     \IEEEauthorblockN{Hongli Zhan}
     \IEEEauthorblockA{\textit{Department of Linguistics} \\
     \textit{The University of Texas at Austin} \\
     \texttt{honglizhan@utexas.edu}
     }
 \linebreakand
     \IEEEauthorblockN{Junyi Jessy Li}
     \IEEEauthorblockA{\textit{Department of Linguistics} \\
     \textit{The University of Texas at Austin} \\
     \texttt{jessy@utexas.edu}
     }
 \and
     \IEEEauthorblockN{Desmond C. Ong}
     \IEEEauthorblockA{\textit{Department of Psychology} \\
     \textit{The University of Texas at Austin} \\
     \texttt{desmond.ong@utexas.edu}
     }
 }

\maketitle

\begin{abstract}

Large Language Models (LLMs) have demonstrated surprising performance on many tasks, including writing supportive messages that display empathy. Here, we had these models generate empathic messages in response to posts describing common life experiences, such as workplace situations, parenting, relationships, and other anxiety- and anger-eliciting situations. Across two studies (N=192, 202), we showed human raters a variety of responses written by several models (GPT4 Turbo, Llama2, and Mistral), and had people rate these responses on how empathic they seemed to be. We found that LLM-generated responses were consistently rated as more empathic than human-written responses. Linguistic analyses also show that these models write in distinct, predictable ``styles", in terms of their use of punctuation, emojis, and certain words. These results highlight the potential of using LLMs to enhance human peer support in contexts where empathy is important.  

\end{abstract}


\section{Introduction}

Artificial Intelligence chatbots have changed so much since the early days of ELIZA \cite{weizenbaum1966eliza} which was modeled after a therapist, and which people found helpful and empathic. In recent years, there have been many successful chatbots (like Wysa \cite{inkster2018empathy} and Woebot \cite{fitzpatrick2017delivering}) that aim to provide empathic responses to help people suffering from mental distress. Indeed, access to social support is vital for personal well-being \cite{cohen1985stress} and the percentage of Americans seeking mental health treatment has almost doubled from 13\% in 2004 to 23\% in 2022 \cite{Brenan2022MentalHealth}.

The recent development of Large Language Models (LLMs) have produced ``general purpose" models, pre-trained in an unsupervised manner on enormous amounts of natural language text, and such models can then generalize to many different tasks that they were not specifically trained on \cite{brown2020language}. 
LLMs like chatGPT, with hundreds of millions of active users, are so successful because they can succeed on so many different tasks, from answering questions to writing reports, or even just having a conversation. 

Researchers have explored using these LLMs to provide social and emotional support \cite{ayers2023comparing, sharma2023cognitive, sorin2023large}. In a recent study, \cite{ayers2023comparing} had people read responses generated by GPT-3 to posts on a healthcare forum (Reddit’s r/askDocs); these responses were rated not only as significantly higher in terms of the quality of medical advice, but also as showing markedly greater empathy, compared to responses written by real physicians. In fact, the effect size was massive---a mean difference of 1.5 on a 5-point scale; which we estimate to correspond to a Cohen’s $d$ of about 1.6 (i.e., 1.6 standard deviations). It is not that physicians are less empathic, but physicians are already overburdened with so many competing demands \footnote{In this case, they were voluntarily responding to inquires on Reddit posts, out of their own personal time}. On the other hand, AI could actually help reduce some of this cost and enhance human capabilities: in a recent experiment, human peer-support providers were paired with AI that gave suggestions to make their responses more empathic; these AI-supported responses were rated as more empathic by those who received them, compared with providers who did not have the AI option \cite{sharma2023cognitive}. In non-peer-reviewed studies, LLM-supported peer support messages also reduced response times and improved quality \cite{germain2023mental}; and others are experimenting with a range of other applications \cite{bannon2023can}, such as using ChatGPT as their personal therapist \cite{khullar2023can} and improving physician–patient communication \cite{kolata2023when}. 

We will caution that these LLMs do not possess ``empathy" as psychologists have defined it (e.g., see also \cite{montemayor2022principle, perry2023ai, inzlicht2023praise}), but people can and do perceive empathy in the text that such LLMs produce. We do not think that AI-displayed empathy should replace, but serve as a complement to, human empathy. But the fact is that human empathy is often in short supply---people are often unable to provide as much empathy as they would like (as it is costly), and there is a large, unmet demand (as evidenced by the success of many peer support and therapy startups: Koko, Talkspace, 7cups). Moreover, beyond scalability, AI-displayed empathy might have other benefits \cite{inzlicht2023praise}. 
For instance, people might also be more open to sharing with AI agents, fearing less judgment and disclosing more sensitive issues \cite{lucas2014s, park2021designing}. 

In all, it is still worth exploring the potential of these models to generate high-quality responses that might be beneficial to some users, and how these responses are perceived. There are downsides to AI-displayed empathy, including appearing insincere and deceptive \cite{montemayor2022principle, cuadra2024illusion}, which we return to discussing more at the end. 

Existing studies on the perceptions of empathy displayed by LLMs have several limitations (See Table \ref{tab:lit_review}). First, they tend to be focused on specific domains, like medical advice \cite{ayers2023comparing, sorin2023large, tu2024towards}. Second, they tend to be focused on specific models (e.g., only GPT \cite{ayers2023comparing} or PaLM \cite{tu2024towards}). Here, we aim to address this gap by studying the perceptions of LLM-displayed empathy across a range of everyday social contexts (e.g., relationship or work-related advice), and a range of models.

\section{Related Work}

\subsection{Human empathy vs AI-displayed empathy}

Human empathy is complex, comprising multiple components, including: \emph{feeling} what someone else is feeling (experience sharing, or ``affective empathy"), being able to understand others' perspectives (perspective-taking or ``cognitive empathy"), and also a motivational component to help alleviate others' distress (empathic concern, or ``motivational empathy") \cite{zaki2012neuroscience, genzer2023empathy}. Based on this definition, many researchers have argued that AI simply cannot \emph{have} empathy in the same manner as humans, as AI do not have emotions or motivations \cite{perry2023ai}; such displays would seem insincere at best, and deceptive at worst \cite{montemayor2022principle, cuadra2024illusion}.

On the other hand, if we take the perspective of a human user, many people do report perceiving empathy in their interactions with AI, whether it be text-based chatbots \cite{weizenbaum1966eliza, inkster2018empathy, fitzpatrick2017delivering} or physical interaction with social robots \cite{darling2015empathic}. People may be perceiving that an AI understands how they are feeling (e.g., supported by NLP work on recognizing emotions \cite{demszky2020goemotions, suresh2021not, suresh2021using} and empathy \cite{rashkin2019towards, sharma2020computational, lee2021building} in text), shows concern, and responds in a contextually-appropriate manner (e.g., \cite{zhou2020design}). Thus, here we avoid the question about whether AI \emph{has} or \emph{can have} empathy (which it does not and can not), but instead focus on people's perceptions of AI-displayed empathy in AI-generated responses (which people do). 

\subsection{LLMs, zero-shot learning, and displayed empathy}

Up through 2020-2021, the dominant paradigm was to take a pre-trained language model, and ``fine-tune" the model on a downstream task like empathy recognition or production (i.e., training a classification layer using a labeled dataset). The most recent ``Large Language Models" (LLMs), with GPT3 \cite{brown2020language} arguably being the first, are primarily defined by their ability to do zero-shot learning. That is, you can give an LLM a natural language description of a new task, and it would solve it (sometimes surprisingly well, sometimes not so). For instance, GPT can identify the emotions and appraisals present in a given piece of text \cite{tak2023gpt, broekens2023fine, zhan-etal-2023-evaluating}.

\begin{table}[htbp]
\caption{Previous papers testing LLMs' empathic response generation}

\begin{center}
    \begin{tabular}{cp{1.6cm}p{3.0cm}}
    \hline
    \textbf{Authors (Year)} & \textbf{\textit{LLMs}} & \textbf{\textit{Domains}} \\
    \hline
    Ayers et al. (2023) \cite{ayers2023comparing} & GPT-3.5 & r/AskDocs (medical advice) \\ 
    \hline
    Tu et al. (2024) \cite{tu2024towards} & PalM-2 &  simulated medical text-based consultations \\ 
    \hline
    Cuadra et al. (2024) \cite{cuadra2024illusion} & PaLM-2, GPT-3.5 Turbo, GPT4 & r/mentalhealth \\
    \hline
    \textbf{The current paper} & GPT4 Turbo, Llama 2, Mistral & 
    r/Anger, 
    r/Anxiety, 
    r/COVID-19\_support
    r/Parenting
    r/relationships
    r/workplace, 
     \\ 
    \hline
    \end{tabular}

\end{center}
\label{tab:lit_review}
\end{table}

One of the initial studies looking at LLM-displayed empathy \cite{ayers2023comparing}, mentioned in the introduction, compared GPT-written responses to posts seeking medical advice on Reddit, with human-written responses. People rated GPT-written responses as being much more helpful (informative) and more empathic than human-written responses. In another study also in the medical domain, \cite{tu2024towards} studied LLM-based ``doctor agents" in simulated text-based consultations, and found that these agents' responses were rated as better on multiple criteria, including empathy, compared to actual primary care physicians. 

In a recent study, \cite{cuadra2024illusion} examined the responses of LLMs to a variety of challenging and sensitive tests, such as expressing empathy towards certain identities (e.g., neurodivergent, religious or anti-religious, homosexual or homophobic). They also found that LLM responses tended to display greater empathy than human responses (using a classifier from \cite{sharma2020computational}).

Finally, some studies have also explored using LLMs to support human users. \cite{sharma2023cognitive} found that responses written by human support providers paired with an AI that gave suggestions to improve their responses, were rated as more empathic. \cite{demszky2023using} discussed using LLMs to support educators' interactions with students. 

In sum, there is much interest in using LLMs to display empathy, as it could, depending on the context, be desirable. And the few studies that exist suggest promising results. However, there is not yet that much research, aside from several studies mainly in the medical and mental health domains. Importantly, these studies often have specific conversational goals where empathy is not the key criteria and we do not have evidence for if these LLMs can display appropriate, contextualized empathy in everyday situations. This paper addresses this gap. 

\section{Experimental studies}

\begin{table*}[htbp]
\caption{Excerpts from Example Posts and Responses by Models (Domain: workplace)}
\begin{center}
\begin{tabularx}{\textwidth}{p{1.4cm}|X}
\hline
\textbf{Type} & \textbf{Content} \\
\hline

Support-seeker & 
\textbf{So I am the youngest one at my job (20sF) My favorite co worker (50sF) is like my ``work mom", she has kids my age and we get a long vey well. I go to her for advice etc, she is awesome. Recently, She broke up with her BF of many many years. She talked about him constantly. They seemed really in love but he broke her heart. She is so sad and I just feel horrible.}

...

\textbf{What is an appropriate way I can cheer her up or show her love? We are close, but still.. coworkers, you know? A gift? Pick up some of her work for her, help her out? I just want her to know she is supported.} \\
\hline
Human & 
Hi! First off, what you are doing right now is so thoughtful and kind. Your coworker is quite lucky to be working with you. In the face of having a heartbroken friend, the most important step to take is to just ``listen" and ``be present." 

...

Being a coworker or not doesn’t and shouldn’t stop you from considering them a friend, and I wouldn’t worry about that fact whenever you are trying to cheer someone up. 
In my opinion, I would go out to grab some grub and catch a movie with them. Maybe don’t directly mention their break up while hanging out with them too. \\

\hline
GPT4 & 
It's really thoughtful of you to want to support your coworker during this tough time. It's clear that you care a lot about her well-being. Given your close relationship, a small, thoughtful gesture could mean a lot. 

...

Sometimes, just being a good listener and offering a shoulder to lean on can be the most powerful gift of all.
Remember, it's important to respect her privacy and not to push for details about the breakup. Just being present, offering a listening ear, and showing kindness can go a long way in helping her feel supported during this difficult time. \\
\hline
GPT4 ``High" & 
I can sense the deep concern and care you have for your coworker, and it's truly touching that you want to support her during this difficult time. It's clear that her well-being means a lot to you, and you're thoughtful in not wanting to overstep boundaries as a coworker. A heartfelt gesture could be very meaningful. 

...

It's important to respect her space and pace, but your genuine offer of support will likely be a comfort to her. Your empathy and willingness to be there for her is a gift in itself. \\
\hline
GPT4 ``Low" & 
Hey, breakups are tough, but she'll get over it. It happens to everyone. Maybe just give her some space and let her deal with it on her own. If you really feel like you need to do something, just buy her a coffee or something small. She's an adult and has probably been through this before, so she'll be fine. Just focus on your work and don't get too involved in her personal life. It's nice you want to help, but there's not much you can do. \\
\hline
Llama2 & 
Oh my goodness, it sounds like your work mom is going through a tough time! \includegraphics[height=1em]{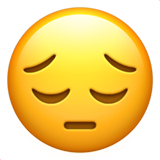} It's so sweet of you to want to support her and help her feel better. \includegraphics[height=1em]{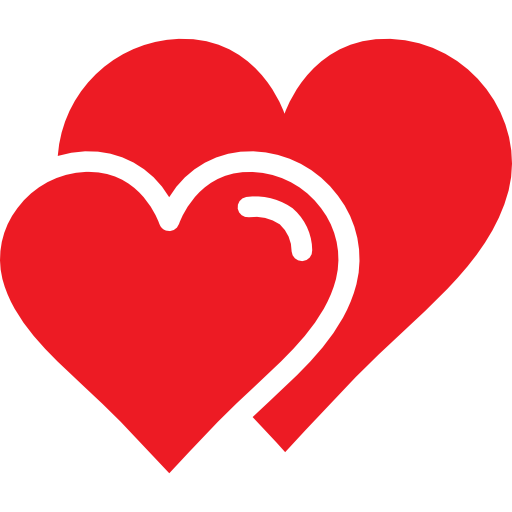}
...

It's okay if she doesn't want to talk about it too much, but it's great that you're there for her when she does need someone to talk to.\includegraphics[height=1em]{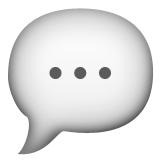}
As for how you can cheer her up or show your support, here are a few ideas:

...

Remember, the most important thing you can do is be there for her and show her that you care.
\includegraphics[height=1em]{emojis/two_hearts.png} 
 She may not be herself right now, but with your support, she'll get through this tough time. \includegraphics[height=1em]{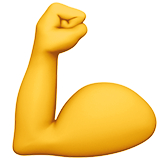} \\

\hline
Mistral & 
I'm so sorry to hear that your work mom is going through a tough time after her breakup, especially during the holidays. It's understandable that you want to support her and help her feel better. You're already doing a great job by being there for her and listening to her when she needs someone to talk to. 

...

Let her know that you're there for her and that she's not alone. Remember, small gestures can go a long way in showing someone that they're loved and supported. \\
\hline

\end{tabularx}
\label{tab:responses}
\end{center}
\end{table*}

We present an overview of our two studies. Both studies were approved (as exempt) by our institution's IRB. 

In Study 1, we compared perceptions of human- and model-displayed empathy, across 3 domains that are relevant to everyday life experiences, namely, Parenting, Relationships, and Workplace. We had a human research assistant, trained in psychology but blind to our hypotheses, craft empathic responses to these situations. This was time-consuming to do well, and so we focused our efforts on a smaller dataset with more comparisons. We chose this design to also address a limitation of previous studies \cite{ayers2023comparing}, where they compared LLM responses with actual human responses from Reddit; these Reddit responses might have been written with the goal to be \emph{informative}, rather than empathic (although \cite{ayers2023comparing} found that LLM responses were rated as more helpful and more empathic than human responses). 

In Study 2, to study the generalizability of our LLM results, we expanded to a larger sample of 120 posts and responses across 6 domains (Anger, Anxiety, COVID-19 support, Parenting, Relationships, and Workplace)---for this second study, we only included LLMs to compare, as human responses tailored to each individual are expensive and effortful. 

\subsection{Study 1 Methods}

\subsubsection{Stimuli} 
We selected posts from three distinct Reddit subreddits: 'r$/$Parenting', 'r$/$relationships', and 'r$/$workplace' (or 'r$/$WorkRant', 'r$/$work'). We chose these domains to sample some of the types of challenges that people may encounter in daily life. We selected a total of 15 posts after manual review, such that the posts are not too sensitive and do not contain private information. Posts varied in length from 70 to 300 words. 

We considered multiple factors to ensure natural and diverse content. Specifically, we aimed to avoid posts that were 1) too short (less than 50 words), 2) too long (more than 400 words), or 3) single questions (such as posts that could be answered with a direct yes/no, or questions lacking substantial content, e.g., `What to do when feeling stressed?'). We selected posts that had some narrative and required more thoughtful and attentive responses. In some instances, this could entail mentioning the perspectives of other individuals, such as the support seeker’s friend, boss, or others who are involved in the situation described in the post. We selected posts that require specific information for a helpful response (e.g., recommending a specific product or remedy to help the support seeker solve the problem at hand).

\paragraph{Human response}

To get high-quality human-written responses, we recruited a research assistant trained in psychology to read these posts and craft natural and empathic responses to these posts. The assistant was instructed to write a 100-150 word response that was contextually appropriate---what would be the best response that would help the person feel better? There were no other restrictions; for example, they could share their own personal stories if they had similar experiences.

\paragraph{Language models}

In our pilot experiments, we experimented with a large set of twenty LLMs (from different model families and configurations within the same family) and different combinations of prompts (e.g., varying length, phrasing, offering definitions). In our tests, smaller models, particularly those under 7 billion parameters, frequently struggled with instructions, exhibiting issues like no response, phrase repetition (e.g., ``It sounds like you you you"), prompt repetition, or irrelevant output.
\footnote{Using the same hyper-parameter setup, we observed that all model responses followed a similar template, making it difficult to discern variations in expressed empathy across models. Therefore, we chose three models that provided consistent responses of similar length to use in our studies.}

After testing, we selected the following three models: \textbf{GPT4 Turbo}, \textbf{LLama2}, and \textbf{Mistral}. 
We used Azure OpenAI API \footnote{https://ai.azure.com} for GPT4 Turbo (\texttt{gpt-4-1106-preview}\footnote{https://platform.openai.com/docs/models/gpt-4-and-gpt-4-turbo}) \- hereafter to referred to as `\textbf{GPT4}'. For Llama2, we used \texttt{Llama2-Chat-13B} version \cite{touvron2023llama}\footnote{https://huggingface.co/meta-llama/Llama-2-13b-chat-hf} and for Mistral, we used \texttt{Mistral (7B) Instruct v0.2}\footnote{https://huggingface.co/mistralai/Mistral-7B-Instruct-v0.2} chat models, using the HuggingFace \footnote{https://huggingface.co/} and Together.AI \footnote{https://together.ai} APIs.

We kept hyperparameter settings similar across all models, setting the maximum token length to 512, temperature to 0.1, top-$p$ (nucleus sampling~\cite{holtzman2019curious}) to 0.9 (for diverse but relevant responses). Special tokens 
were adjusted according to each model's specific template requirements ($[$INST$]$, $[/$INST$]$, or $<$s$>$, $</$s$>$). In the process of developing the prompts, 
we ensured that minor changes (e.g., missing comma, syntax change) did not lead to significant alterations in the output content or errors (e.g., unintended repetitions of a single word or sentence) by generating each model's output multiple times (approx. 2 to 10) for consistency. 

\paragraph{Prompt conditions}

LLMs can be ``prompted" to answer complex queries by instructing them in natural lanugage.This had led to a class of techniques called \emph{prompt engineering} that aim to figure out the best components of a prompt, to yield better results. In our two studies, we adopted several prompting ``best practices"\footnote{https://platform.openai.com/docs/guides/prompt-engineering},
as we describe below, but there is still much room to explore in terms of prompt engineering (we return to this in the Discussion). 

For the `Base' prompt that we used across all the models (Fig. \ref{fig:prompt}), we included the following components: a `role' for the LLM (``you are a peer supporter"), a specific task/goal (``write empathic response"), and a specific format (``limit response to minimum 100 to maximum 150 words").

We additionally explored whether a model can be prompted to display more or less empathy. We designed additional prompts to elicit a ``high" amount of empathy and a ``low" amount of empathy (``generate responses according to levels of empathic accuracy"). In our initial piloting, we also experimented with providing the model with a definition of \textit{empathic accuracy}, as ``the ability to understand the emotion and situation of the other person"; the model's responses did not differ in quality with or without the definition. 
In our testing, we also found that GPT4 was the most compliant with instructions (on length/style) and so we decided to focus only on GPT4 for this ``high" / ``low" empathy \footnote{The ``low" empathy responses, in particular, were also helpful in breaking up the repetitiveness of the empathy rating task, and to ensure that raters were not just giving the same rating to all LLM responses.} conditions. In total, we had 5 model conditions. 

\begin{figure}[tbh]
    \centering
\begin{tcolorbox}[colback=gray!10!white,colframe=blue!50!black,
  colbacktitle=blue!40!black,title=System Instruction Prompts]
  
    \textbf{Base conditions (all models)}\\
    ``You are a peer supporter. Read the support seekers' post and write appropriate and empathic response. Limit your response minimum 100 words to maximum 150 words. Do not exceed 150 words." \\[2ex]
    \textbf{Empathy level conditions (GPT4)}\\
    ``You are a peer supporter. Read the support seeker's post and write responses at different levels of empathic understanding. Draft separate responses that would correspond to each level on a very low and very high level in terms of empathic accuracy. Each response should reflect the level of understanding. Limit each response to a minimum of 100 words and a maximum of 150 words. Do not exceed 150 words."
    \end{tcolorbox}
    
    \caption{Base prompt (Study 1 and 2; all models) and ``Empathy level" prompts (Study 1, GPT4-only) 
    }
    \label{fig:prompt}
\end{figure}

\subsubsection{Participants and procedures} We recruited 200 participants on Prolific to rate human or LLM responses. Seven raters were excluded for taking too long or too short, for a final sample of 193 raters (mean age = 42.6 (SD=14.1), 50.3\% female, 2.6\% non-binary/did not disclose). 
Raters read a post by a support seeker, followed by a response, and were asked to rate on a 5-point Likert scale of (i) how appropriate was the response (with 1 being \textit{extremely inappropriate} and 5 being \textit{extremely appropriate}), and (ii) how empathic was the response (with 1 being \textit{not empathic} and 5 being \textit{very empathic}). 
On subsequent pages, they would see the same post, with a different response, until they have rated all six responses (i.e., models were completely within-subjects; every participant saw every response to a post). Responses were randomized with no indication whether they were AI-generated. We had 15 total posts; one group of raters responded to 7, while a second group responded to 8; the average time taken for raters were about 20 to 25 minutes per person. 

\begin{figure}[tbh]
    \centering
    \includegraphics[width=\columnwidth]{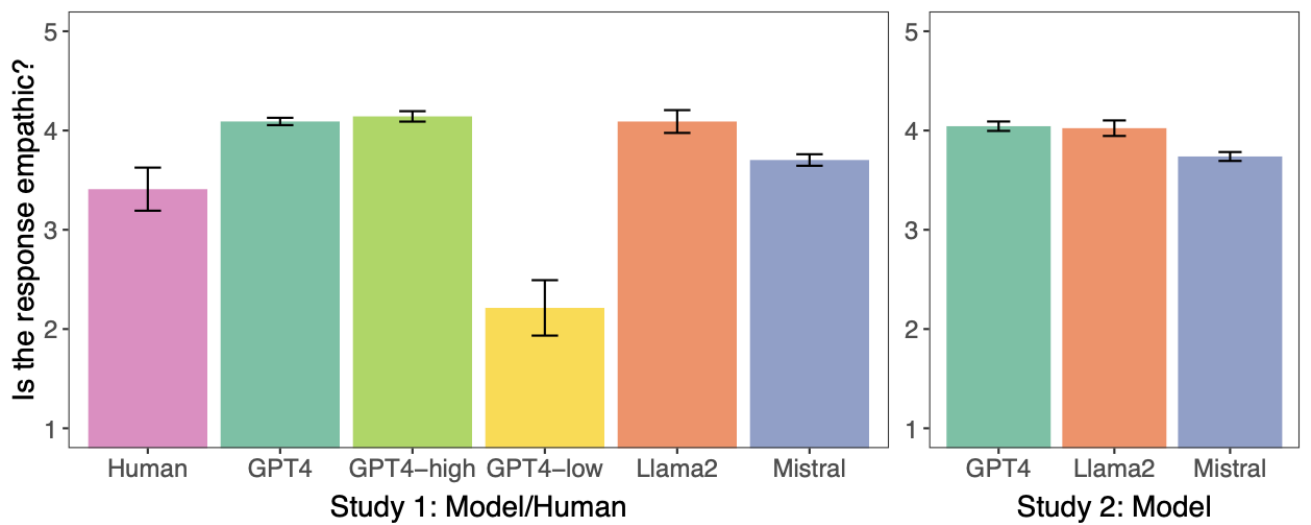}
    \caption{Results of Study 1 (Left) and Study 2 (Right). Mean empathy ratings with 95\% Confidence Intervals, calculated across posts.} 
    \label{fig:study_results}
\end{figure}

\subsection{Study 1 Results}

To statistically analyze our repeated measures data, we fit a mixed-effects linear model predicting our dependent variable (empathy) using whether the response was written by model/human as a (six-level) categorical variable. We also added random effects by raters and by post. 

On average, our human-written responses were judged to be appropriate ($M_\text{human}$ = $3.99$, 95\% CI = $[3.77, 4.21]$, on a 1-5 scale) and empathic ($M_\text{human}$ = $3.41$ $[3.19, 3.62]$). But the human-written responses were rated as less empathic compared to all the models: GPT4 ($M_\text{GPT4}$ = $4.09$ $[4.05,	4.13]$, $t$ = $20.13$, $p<.001$), Llama2 ($M_\text{Llama2}$ = $4.09$ $[3.98, 4.21]$, $t$ = $20.13$, $p<.001$), and Mistral ($M_\text{Mistral}$ = $3.70$ $[3.65, 3.76]$, $t$ = $8.68$, $p<.001$). We found the same pattern of results for response appropriateness.

As mentioned earlier, we included a ``GPT4-high" (GPT4 prompted to respond with high levels of empathy) and a ``GPT4-low" (low levels of empathy) comparison condition. As expected (and as a sanity check), the mean empathy for GPT4-low was by far the lowest across all conditions ($M_\text{GPT4-low}$ = $2.21$ $[1.93, 2.49]$), and was significantly different from all the other conditions. 
We found no statistically significant difference in empathy between GPT4-high and base-GPT conditions ($M_\text{GPT4-high}$ = $4.14$ $[4.09, 4.20]$, $t$ = $1.51$, $p$ = $0.13$). This suggests that GPT4, when asked to write an empathic response without any additional instructions, might already be generating a high amount of empathy. 

Although we did not have any hypotheses comparing GPT4, Llama2, and Mistral, we found that responses written by GPT4 and Llama2 were not significantly different in terms of empathy (in fact, their means were coincidentally exactly the same in our sample $t$ = $0.00$, $p=1$), although both were rated as more empathic than Mistral (both $t$ = $11.45$, $p<.001$).

\subsection{Study 1 Discussion}

The results of Study 1 corroborated with the results of \cite{ayers2023comparing}. However, there is one major difference between the two studies: \cite{ayers2023comparing} compared LLM responses with actual human responses from medical professionals on `r/askDocs' who may not have had the goal to be empathic or may have been too busy \footnote{Although \cite{ayers2023comparing} also found that LLM responses were still rated as higher-quality and more empathic even when compared with just the longest physician responses}.
On the other hand, our study compared LLM responses with responses from a human research assistant trained in psychology, blind to our hypotheses, but told explicitly to display empathy. Even with explicit instruction to display empathy, we found a significant difference between human and LLM responses. 

One limitation of Study 1 was the small sample size of posts (15 posts across 3 domains (parenting, relationships, workplace) as producing human-written responses was time-consuming. In Study 2, we focused on LLM responses, and extended to a larger sample of 120 posts across 6 domains to test if our LLM results generalize across more domains. 

\subsection{Study 2 Methods}

We selected 120 Reddit posts from six domains: Anger (r/Anger), Anxiety (r/Anxiety), COVID-19 related situations (r/COVID-19\_support), Parenting (r/Parenting), Relationships (r/Relationships), and Workplace (mainly from r/workplace, but also used relevant posts from r/Work, or r/WorkRant). Posts were on average 141 words (SD=73.22). 
The rest of the study design was similar to Study 1, except without human-written responses or the high/low empathy prompts. We used the same ``base" prompt as in Study 1 to generate responses for GPT4-Turbo, Llama2-13b-chat, and Mistral-7b-Instruct-V0.2. 
Raters saw ten posts (drawn from the pool of 120) with three responses per post, rating them in the same manner as in Study 1. 
We recruited a total of 203 participants on Prolific to rate these statements. One rater was excluded for taking too long, for a final sample of 202 raters (mean age = 39.9 (SD=13.2), 52.5\% female, 3.5\% non-binary/did not disclose). 

\subsubsection{Linguistic analyses}
With the larger amount of language data in Study 2, we conducted additional linguistic analyses to characterize the LLM responses. 

\paragraph{LIWC}
We used the LIWC (Linguistic Inquiry and Word Count; \cite{pennebaker2007linguistic}) to calculate additional linguistic features. 

The latest version, LIWC-22, offers up to 80 categories of language dimensions, which are supported by hundreds of psychology studies that demonstrated correlations between language use and psychological characteristics (e.g., personality, mental states, anxiety) 
of the writer  \cite{tausczik2010psychological,sonnenschein2018linguistic, kleinberg2020measuring, shen2017detecting}. In our analysis, we do not seek to anthropomorphize or make any claims about the ``psychological processes" ``inside" the models: we mainly use these dictionaries to characterize the types of language that these models are using.

We were interested in three categories of features. For each of these categories, we calculated the frequency of use (i.e., word count), normalized by the total number of words to give a proportion. The first category is the use of different types of \textbf{pronouns}: first-person singular (``I") and plural (``We"), second-person (``You") and third-person singular (``he/she") and plural (``they"). Second, we looked at the use of \textbf{punctuation}, specifically, question marks, exclamation marks, and emojis. Third, we were interested in \textbf{emotion-related words}---LIWC calculates separate scores for positive emotion words and for negative emotion words (e.g., anxiety-, anger-, and sadness-related words). 

\paragraph{Bag-of-words features predicting model identity and empathy level}
To see if different models tend to generate different styles of responses, we calculated bag-of-words features (i.e., $n$-gram features), and examined if these features are predictive of which model generated the response. We trained a multi-class logistic regression to predict which LLM produced a given response, using the top $200$ most-frequent unigrams and bigrams, and the $20$ most-frequent emojis. We used a $90$:$10$ train-test partition. To pre-process the text, we excluded stopwords and punctuations, before extracting the $n$-grams. We normalized the features, and performed L2 regularization.

Lastly, we used a regression model to predict the empathy ratings using the same bag-of-words features.

\subsection{Study 2 Results}

\subsubsection{Statistical analysis}

Our results from Study 2 replicated our results from Study 1 with a larger set of 120 posts. On average, GPT4 responses were rated as empathic ($M_\text{GPT4}$ = $4.04$ $[4.00, 4.09]$), as were Llama2 responses ($M_\text{Llama2}$ = $4.02$ $[3.95, 4.10]$), about 4 out of 5, and these were not significantly different from each other ($p$ = $.41$). Mistral responses were rated a little less empathic ($M_\text{Mistral}$ = $3.74$ $[3.69, 3.78]$), and this was significantly less than the other two models (Mistral $<$ GPT, $t$ = $12.6$, $p<.001$; Mistral $<$ Llama2, $t$ = $11.8$, $p<.001$). Comparing across the results from Study 1 and 2, there were also no differences in the empathy ratings across both studies for GPT ($p=.59$), Llama2 ($p=.58$), and Mistral ($p=.74$), suggesting that the samples between two studies are equivalent.

\subsubsection{Linguistic Analyses}

\begin{figure}
    \centering
    \includegraphics[width=.85\linewidth]{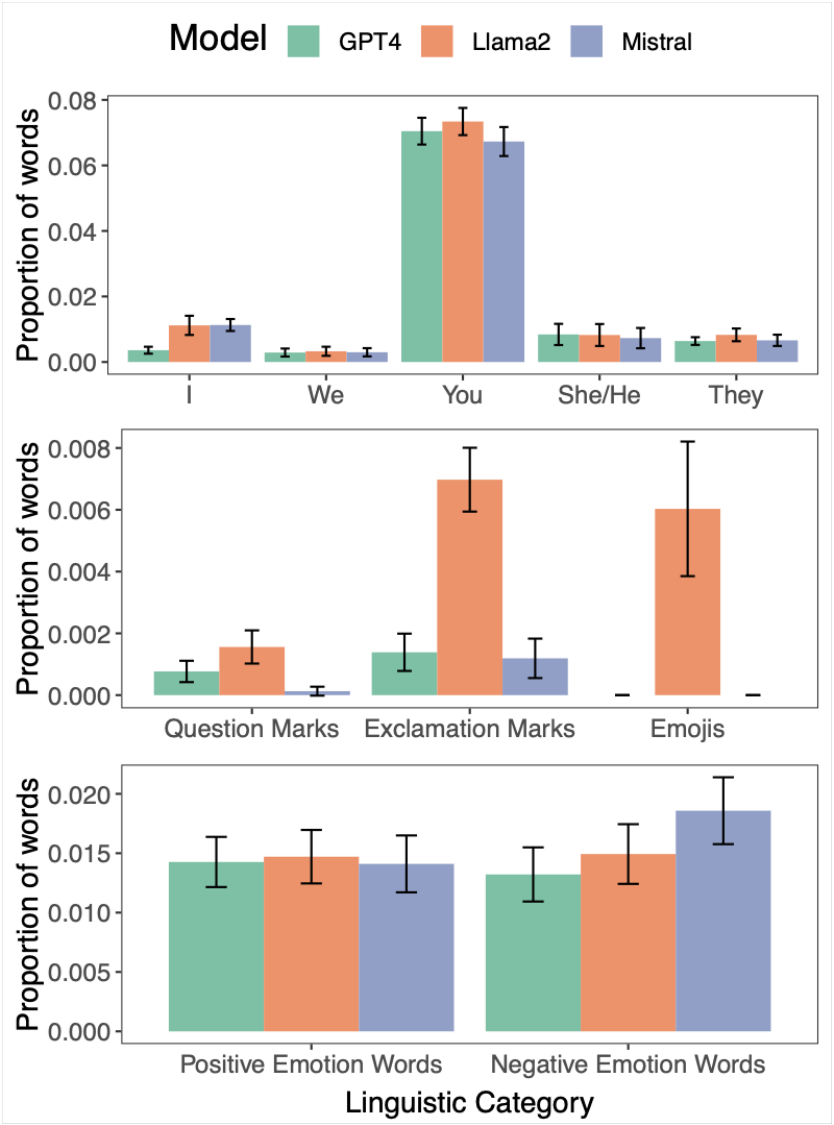}
    \caption{Study 2: Results from LIWC Analyses. Top: Pronoun frequency, Middle: Punctuation, Bottom: Emotion words 
    }
    \label{fig:liwc_plots}
\end{figure}

Compared to the other models, Llama2 responses were the most verbose ($M_\text{Llama2}$ = $219$ words, SD$_\text{Llama2}$ = $62.2$), followed by GPT4 ($M_\text{GPT4}$ = $186$, SD$_\text{GPT4}$ = $28.5$), and Mistral ($M_\text{Mistral}$ = $137$, SD$_\text{Mistral}$ = $34.2$), noting that our prompt instructed models to keep between 100 and 150 words.

There were differences in pronoun frequency (Fig. \ref{fig:liwc_plots}). GPT4 responses contained less ``I" words compared to Llama2 ($b$ = $.0076$, $t$ = $6.32$, $p<.001$) or Mistral ($b$ = $.0077$, $t$ = $6.42$, $p<.001$) responses. By contrast, Llama2 responses contained more ``You" words than Mistral ($b$ = $.0061$, $t$ = $2.95$, $p$ = $.004$), and GPT4 (although this difference was not significant, $p$ = $.16$) responses.
There were no significant differences in the frequency of ``We" ($p$'s$>.65$), ``She/he" ($p$'s$>.25$), and ``They" ($p$'s$>.07$) words. .

For punctuation, there were also large differences: LLama2 responses contained more question marks (asking more questions, e.g., ``Have you considered talking to someone about how you're feeling?") compared to GPT ($b$ = $.0008$, $t$ = $3.10$, $p$ = $.002$) or to Mistral ($b$ = $.0014$, $t$ = $5.6$, $p<.001$). 

Llama2 responses also contained significantly more exclamation marks than GPT4 ($b$ = $.0056$, $t$ = $12.6$, $p<.001$) and Mistral ($b$ = $.0058$, $t$ = $13.0$, $p<.001$). And perhaps most strikingly, neither GPT4 nor Mistral responses contained any Emojis, but Llama2 responses had frequent use of Emojis.

Finally, we examined the frequency of emotion words. The frequency of positive emotion words did not differ across model responses ($p$'s$>.54$). For negative emotion words, Mistral responses contained the greatest frequencies of negative emotions compared to Llama2 ($b$ = $.0036$, $t$ = $3.75$, $p<.001$) and GPT4 ($b$ = $.0054$, $t$ = $5.52$, $p<.001$) responses. The difference in the frequency of negative emotion words between Llama2 and GPT4 responses was not significant ($p=.08$).

\begin{table}[tbh]
  \small
  \centering
  \renewcommand{\arraystretch}{1.2}
  \caption{Results of the logistic model trained to predict which LLM generated a response. Top: F1-scores. Bottom: Top $20$ features (unigrams, bigrams, emojis) with the most positive weights.}

  \adjustbox{max width=\linewidth}{
    \begin{tabular}{C{2cm}S[table-format=1.2,table-align-text-post=false]||C{1.4cm}S[table-format=1.2,table-align-text-post=false]||C{1.5cm}S[table-format=1.2,table-align-text-post=false]}
    \toprule
        \multicolumn{2}{c||}{\textsc{GPT4}} & \multicolumn{2}{c||}{\textsc{Llama2}} & \multicolumn{2}{c}{\textsc{Mistral}} \\
        \emph{F1 scores:} & 0.92 & & 0.96 & & 0.86 \\
        
        \textbf{Feature} & \textbf{Weight} & \textbf{Feature} & \textbf{Weight} & \textbf{Feature} & \textbf{Weight} \\
    \midrule
        completely & 1.23  & hey   & 1.66  & sorry hear & 1.74 \\
        feelings & 0.99  & oh    & 1.37  & sorry & 1.65 \\
        completely understandable & 0.93  & totally & 1.18  & important remember & 1.20 \\
        seeking & 0.86  & like  & 1.09  & hear  & 1.02 \\
        professional & 0.86  & sounds like & 1.01  & really sorry & 0.93 \\
        seek  & 0.81  & needs & 0.94  & feeling & 0.90 \\
        approach & 0.80  & sounds & 0.93  & understand & 0.88 \\
        health & 0.78  & helpful & 0.92  & difficult & 0.82 \\
        friends & 0.76  & talk  & 0.84  & consider & 0.79 \\
        positive & 0.73  & normal & 0.79  & ask   & 0.56 \\
        incredibly & 0.72  & want  & 0.74  & experience & 0.56 \\
        strategies & 0.72  & \includegraphics[height=1em]{emojis/two_hearts.png} & 0.72  & remember & 0.55 \\
        especially & 0.68  & great & 0.71  & unique & 0.54 \\
        life  & 0.68  & possible & 0.68  & things & 0.54 \\
        understandable feel & 0.67  & important & 0.66  & potential & 0.48 \\
        situation & 0.66  & prioritize & 0.65  & mental & 0.43 \\
        new   & 0.66  & overwhelmed & 0.64  & prioritize & 0.43 \\
        challenging & 0.64  & try   & 0.64  & long  & 0.41 \\
        natural & 0.63  & don   & 0.60  & really & 0.40 \\
        experiencing & 0.61  & normal feel & 0.58  & natural & 0.39 \\
    \bottomrule
    \end{tabular}}
    \label{tab:study2_logistic_regression}
\end{table}

\begin{table}[htbp]
  \centering
  \caption{Weights in OLS regression predicting empathy ratings.}
  \label{tab:OLS_top_weights}
    \begin{tabular}{c|S[table-format=1.2,table-align-text-post=false]||c|S[table-format=1.2,table-align-text-post=false]}
    \toprule
        \multicolumn{1}{c|}{\textbf{Feature}} & \multicolumn{1}{c||}{\textbf{Weight}} &
        \multicolumn{1}{c|}{\textbf{Feature}} & \multicolumn{1}{c}{\textbf{Weight}} \\
    \midrule
        \includegraphics[height=1em]{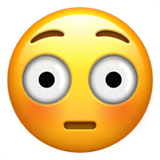}     & 17.59 &
          moment &	1.83$^{**}$ \\
          
        \includegraphics[height=1em]{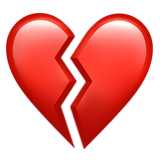} & 12.40$^*$ & valid &	1.81$^*$ \\
        
        \includegraphics[height=1em]{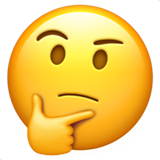}    & 9.86 & 
          relationship & 1.78$^{**}$ \\

        \includegraphics[height=1em]{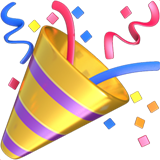}    & 7.01 &
          incredibly & 	1.76 \\

        \includegraphics[height=1em]{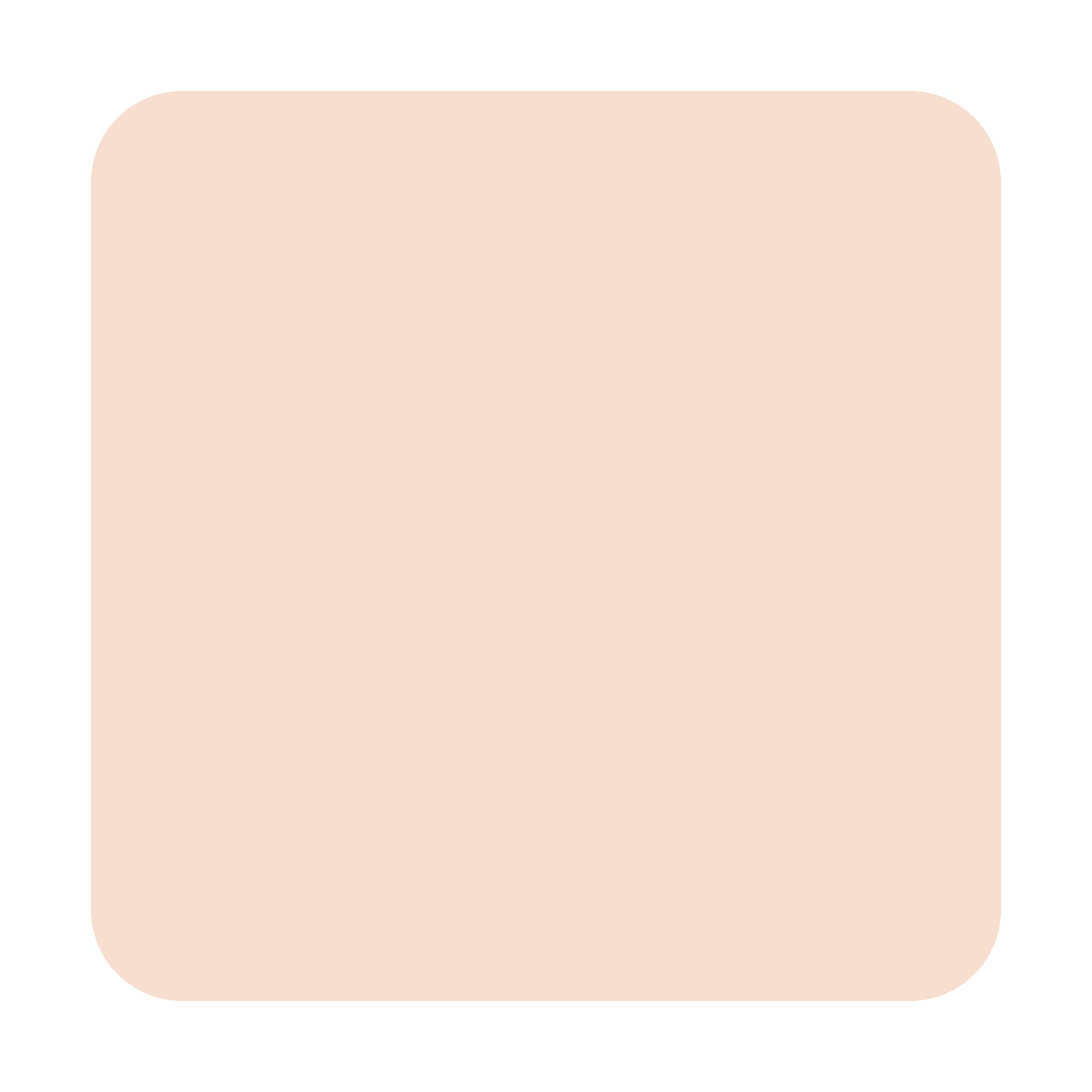}    & 5.10 &
          going &	1.70$^*$ \\

        \includegraphics[height=1em]{emojis/pensive.png}    & 4.32 & 
          step   & 1.63$^*$ \\

        \includegraphics[height=1em]{emojis/muscle.png}    & 4.26 &
          environment	 &  1.63$^*$ \\

        sorry & 3.66$^*$ &	thing & 1.56 \\

        really sorry & 2.72$^*$ & understanding & 1.50$^*$ \\

        day	& 2.03$^{**}$ & \includegraphics[height=1em]{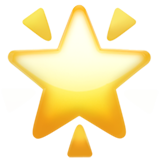}    & 1.46 \\
    \bottomrule
    \multicolumn{4}{l}{\textsuperscript{*}$p<0.05$, \textsuperscript{**}$p<0.01$, \textsuperscript{***}$p<0.001$}
    \end{tabular}
\end{table}

Next, we examined whether the LLM responses have their own distinctive styles, by training a logistic regression model to predict which LLM generated a given response. Our model does well (see Table \ref{tab:study2_logistic_regression}, achieving an F1 score of $.92$ in classifying GPT4, $.96$ in classifying LLama2, and $.86$ in classifying Mistral. If we examine the most predictive weights, we can see that GPT4 responses tend to contain words like ``completely understandable", and seem to recommend additional resources for the author (e.g., ``seek/seeking", ``professional", ``friends", ``strategies"). LLama2 responses, compared to the other two models, is characterized by a more casual manner, with the five most predictive features being: ``hey", ``oh", ``totally", ``like", and ``sounds like". Mistral responses, compared to the other two models, contains more apologies, with ``sorry [to] hear" (to is a removed stop-word), ``sorry", ``really sorry", being features with high discriminability. 

Finally, we used these features to predict the empathy ratings given by raters (Table \ref{tab:OLS_top_weights}). Our model does very well, with a Mean Absolute Error of $0.44$ on a 5-point Likert scale. The features with the most positive weights predicting empathy ratings seem to be mostly emojis (although only the ``broken heart" emoji is statistically significant), and apologies (``really sorry", ``sorry").

\subsection{Study 2 Discussion}
With a larger sample of posts, the results of Study 2 were consistent with what we observed in Study 1: On average, responses by the LLMs were perceived as highly empathic. 
Although the prompts contained instructions to restrict responses to between $100$ to $150$ words, LLM responses tended to be longer. This could be because people tend to prefer longer model responses \cite{singhal2023long}, which have been factored into their training.  
Our linguistic analyses of the responses revealed distinct characteristics among the responses generated by different LLMs. For instance, Llama2 responses tend to be more casual (``hey", ``totally") and containing more exclamation and question marks, and emojis. By contrast, Mistral responses tend to contain more condolences (``sorry [to] hear", etc) and acknowledgement of the negative emotions in the situation. 

\section{Discussion}

Across two studies, we investigated whether current LLMs can produce appropriate and empathic responses to support-seekers experiencing various everyday situations.  
In Study 1, we found that LLM-generated responses were rated as being more empathic than human-written responses, and that specifically prompting GPT to be more empathic did not significantly increase ratings of perceived empathy.

In Study 2, we examined LLM-generated responses to a larger sample of posts across more domains, and replicated our results that LLM-produced responses were perceived as highly empathic. Exploratory linguistic analyses showed that there were different characteristic patterns in model responses which leads to high accuracies in classifying these models. For instance, Llama2 responses were the most ``casual".

Our results suggest that LLMs, trained on a large amount of text data that presumably contain similar support-seeking posts and responses, are able to produce responses that are perceived to be empathic. There are still many challenges. LLMs are limited in many ways: for example, they may not have access to much of the context that a human-human interaction might entail---there are no non-linguistic cues available to text-only LLMs. Moreover, we only looked at single-turn responses; over the course of a conversation, LLM responses would need other elements, such as long-term memory, and keeping track of the current conversational state (including a user's current emotions). 

There are also broader, ethical issues. Many researchers have rightly pointed out the inherent deception in AI agents displaying empathy \cite{montemayor2022principle, cuadra2024illusion}, many of which parallel arguments against AI agents having or displaying ``emotions" \cite{cowie2012good, devillers2023ethical}. As mentioned earlier, we do not think LLMs can or should replace human empathy, but there are many other possibilities that are still worth exploring such as how LLMs might be able to augment human-human connections \cite{sharma2023cognitive, demszky2023using, tu2024towards}, or provide empathic responses to people who opt-into it (e.g., using a chatbot app). 

In sum, our results suggest a promising start, and also point to many future questions. For instance, the LLMs in our studies have markedly different ``styles" of displaying empathy. Could we measure people's preferences for these ``empathic styles", and use prompting or other methods to tailor responses to people's preferences? We believe that, used carefully, LLM-generated empathic responses could be beneficial to people's well-being.

\section*{Ethical Impact Statement}
The studies were approved by our institution's IRB. We used public data (Reddit) and additionally manually verified that each post was anonymized and does not contain any identifiable information. 

There are still many risks associated with using LLMs. First, they may ``hallucinate" and produce false information (which, if given in the context of e.g., medical advice, could be dangerous). Second, there is no guarantee that LLMs will not generate harmful content, even in a context where one is asking it to display empathy (e.g., \cite{cuadra2024illusion}). Third, LLM responses may also contain biases and beliefs (e.g., cultural, political) which might overrepresent certain groups \cite{santurkar2023whose, tao2023auditing}. 

There are also potential mis-uses of this type of research, such as to manipulate people's emotions. However, many commercial companies that offer LLMs, such as OpenAI's GPT4, do have safety guardrails in place, such that these models may refuse to generate content discussing certain topics. These are not perfect, but are one approach to mitigating such harms. However, there are no guardrails for open-sourced LLMs.

\section*{Acknowledgment}
 We thank our research assistants, Ryan Truong and Emily Yang for their assistance in data collection. We would like to thank Judith Amores, Mary Czerwinski, Javier Hernandez, Lindy Le, and Gonzalo Ramos (in alphabetical order by last name) from the Microsoft Research Human Understanding and Empathy (HUE) group for their invaluable support and feedback throughout the research process.

\section{Resources}

We release our dataset at
\url{https://github.com/yoonlee78/LLM_empathy_social_support.git}

\bibliographystyle{IEEEtran}
\bibliography{biblio}

\end{document}